\documentclass[10pt, a4paper]{article}
\usepackage{lrec2022} 
\usepackage{multibib}
\newcites{languageresource}{Language Resources}
\usepackage{graphicx}
\usepackage{tabularx}
\usepackage{soul}
\usepackage{amssymb}
\usepackage[T1]{fontenc}

\usepackage{titlesec}
\titleformat{\section}{\normalfont\large\bfseries\center}{\thesection.}{1em}{}
\titleformat{\subsection}{\normalfont\SmallTitleFont\bfseries\raggedright}{\thesubsection.}{1em}{}
\titleformat{\subsubsection}{\normalfont\normalsize\bfseries\raggedright}{\thesubsubsection.}{1em}{}
\renewcommand\thesection{\arabic{section}}
\renewcommand\thesubsection{\thesection.\arabic{subsection}}
\renewcommand\thesubsubsection{\thesubsection.\arabic{subsubsection}}

\usepackage{epstopdf}
\usepackage[utf8]{inputenc}

\usepackage{hyperref}
\usepackage{xstring}

\usepackage{color}

\title{Guidelines and a Corpus for Extracting Biographical Events\\ \vspace*{.5\baselineskip}}

\name{Marco A. Stranisci$^{\ast}$, Enrico Mensa$^{\ast}$, Ousmane Diakite$^{\ast}$, \\
\bf \large Daniele P.  Radicioni$^{\ast}$, Rossana Damiano$^{\ast}$} 

\address{$^{\ast}$Department of Computer Science - University of Turin, Turin, Italy \\
{\tt \small \{marcoantonio.stranisci, enrico.mensa, daniele.radicioni, rossana.damiano\}@unito.it}\\
\tt \small ousmane.diakite@edu.unito.it}

\abstract{
Despite biographies are widely spread within the Semantic Web, resources and approaches to automatically extract biographical events are limited. Such limitation reduces the amount of structured, machine-readable biographical information, especially about people belonging to underrepresented groups. Our work challenges this limitation by providing a set of guidelines for the semantic annotation of life events. The guidelines are designed to be interoperable with existing ISO-standards for semantic annotation: ISO-TimeML (SO-24617-1), and SemAF (ISO-24617-4). Guidelines were tested through an annotation task of Wikipedia biographies of underrepresented writers, namely authors born in non-Western countries, migrants, or belonging to ethnic minorities. $1,000$ sentences were annotated by $4$ annotators with an average Inter-Annotator Agreement of $0.825$. The resulting corpus was mapped on OntoNotes. Such mapping allowed to to expand our corpus, showing that already existing resources may be exploited for the biographical event extraction task.  
 \\ \newline  }

\begin{document}

\maketitleabstract

\section{Introduction}
The Semantic Web shift led in few years to a growth of biographical information online. Knowledge Graphs (KG), such as Dbpedia \cite{auer2007dbpedia} and Wikidata \cite{vrandevcic2014wikidata}, allow the gathering of structured socio-demographic attributes and facts about people. Notwithstanding, many unstructured data conveying biographical information are still not mapped in KGs. Wikipedia pages express more content than their corresponding Wikidata profile: for instance, all the places where a person lived within their life and all their migrations. The enrichment of existing KGs with such information would be crucial in improving several tasks such as community detection \cite{wang2018acekg}, prosopography \cite{booth2008orlando}, and social bias detection \cite{sun2021men}.

Although several semantic models have been proposed to formally represent a biographical event \cite{krieger2015owl,tuominen2018bio}, computational resources for the automatic extraction of biographical events from text are still missing, and there are no annotated corpora, nor annotation schemes specifically designed for this task.  

In this paper, we describe a novel set of annotation guidelines specifically developed for this task, built on two Semantic Annotation Frameworks, ISO 24617-1 \cite{pustejovsky2010iso}, and ISO 24617-4 \cite{bunt2013conceptual}. The guidelines have been adopted to annotate a corpus of $1,000$ sentences extracted from Wikipedia pages of under-represented writers, namely writers born in non-Western countries, migrants or belonging to ethnic minorities \cite{stranisci2021representing}. The resource is designed to be interoperable with existing language resources \citelanguageresource{pustejovsky2003timebank,hovy2006ontonotes}, in order to augment the corpus with additional data through a systematic mapping. Such data augmentation is crucial for the future implementation of a pipeline for the automatic extraction of biographical events. 

The paper is structured as follow. In Section \ref{sec:related}, a review of works on biographical encoding and event extraction is provided. Section \ref{sec:data_gathering} describes data collection and annotation guidelines design. In Section \ref{sec:results}, results of the annotation are presented. Section \ref{sec:mapping} presents the mapping of the resource with existing corpora. Finally, Section \ref{sec:conclusions} concludes the paper with some insights on future work.

\section{Related Work}\label{sec:related}

The extraction of biographical events from text brings into play two main research lines, namely... 

\paragraph{Semantic Roles and Events Annotation Frameworks.} The annotation of semantic roles has been addressed by a number of approaches with specific focuses (see  \newcite{petukhova-bunt-2008-lirics}). 
FrameNet (FN) \cite{baker1998berkeley} and PropBank (PB) \cite{kingsbury2002treebank} are two databases of semantic roles: the former is not syntactically bounded and relies on a detailed taxonomy of semantic roles; the latter is centered on verbs and the classification of arguments is coarse-grained.
Other approaches are focused on a general notion of semantic role. VerbNet's (VN) \cite{schuler2005verbnet} aim is the classification of English verbs on the basis of semantic-syntactic properties; LIRICS identifies `relational notions which
link a participant to some real or imagined situation (`event')' \cite{bunt2002requirements}. In last years, attempts to unify such resources have been made. The Semantic Annotation Framework (SemAF) \cite{bunt2013conceptual} provides an unifying framework according to which a semantic annotation relies on a finite set of eventualities (EV) and participants (PT) that form entity structure pairs with \textit{markables}, namely tokens to which an EV or PT can be attached. Pairs are then combined in links through link structure. For instance, in the sentence `she published poetry' three entity structure pairs may be annotated: $\epsilon_1$ = $<$ She, POET $>$; $\epsilon_2$ = $<$ published, PUBLISH $>$, and $\epsilon_3$ = $<$ poetry, POEM $>$. A link structure triple connect $\epsilon_1$ and $\epsilon_2$, assigning to the former the role of agent: $L_1$ = $<$ $\epsilon_1$, $\epsilon_1$, Agent$ >$.

Frameworks for the annotation of events are heterogeneous, reflecting the high variety of existing event extraction tasks (see  \newcite{xiang2019survey}). The ACE/ERE initiative \cite{song2015light} resulted in a series of news corpora in which textual triggers were annotated and labelled by referring to a close set of event types. For instance, the word `migration' triggers an event of the type `Movement'. The Topic Detection and Tracking initiative (TDT) \cite{allan2012topic} led to a corpus in which the story rather is annotated and labelled with reference to actual historical events (eg: Death of Kim Jong II, Cuban Riot in Panama, etc.) rather than general categories. The ISO-TimeML framework \cite{pustejovsky2010iso} is a standard for the annotation of temporal expressions, events, and temporal relations between events. According to such approach, an instance of the type `EVENT' must be used to annotate a situation that happens or occurs. Furthermore, events are categorized by some linguistic properties. For instance, the word `start' triggers an event of the type `ASPECTUAL', whereas `say' is a `REPORTING' event. The Richer Event Description (RED) guidelines \cite{o2016richer} is a reformulation of ISO-TimeML in which the taxonomy of event properties is simplified, but further annotation layers are defined: entities, causal relations between events, and link between entities.

Our annotation guidelines for biographies take inspiration from two existing frameworks. On one side, they adopt the semantic formalism of SemAF \cite{bunt2013conceptual}, while on the other side they partly inherit the taxonomy of events proposed in ISO-TimeML \cite{pustejovsky2010iso}.

\paragraph{Biographical Events Extraction.} Despite the existence of several semantic models for biographical events encoding, few works focused on the extraction of biographical information. Russo et al.~\cite{russo2015extracting} collected $782$ biographies of people deported to Nazi concentration camps with the aim of extracting a predefined set of information  from both raw text and DBpedia. Then, all information was arranged into a structured representation by using the TimeML framework~\cite{pustejovsky2010iso}. \newcite{menini2017ramble} defined a set of verbal motion frames and used it to extract migration events from Wikipedia biographies. 

Both works adopt a top-down approach. First, a number of information to be retrieved is defined, then an event extraction pipeline is built. 

%
%
Our guidelines rely on a bottom-up approach: instead of a predetermined classification of event types to be extracted, the focus is on all events in which the entity of the type \textit{writer} is involved as a participant.

\section{Data Collection and Annotation Scheme Design}\label{sec:data_gathering}
In this section, the data gathering and preprocessing from Wikipedia is described; then, the annotation guidelines are presented.

\subsection{Data Gathering}
The corpus is a collection of sentences extracted from $8,047$ Wikipedia English pages of under-represented writer, namely authors born in non-Western countries, migrants or ethnic minorities. Specifically, Wikidata properties `place of birth', `occupation', and `ethnic group' were exploited in order to identify all writers born in a former colony or writers belonging to a minority group that were born in a Western country. The data gathering process was performed in four steps: (i) each biography was split in sentences using Stanford Core NLP \cite{manning2014stanford}; (ii) for each sentence, all the named entities of the type Location or Organization were identified using the same tool; (iii) an automatic semantic role labelling was performed on each sentence, using SRL Bert \cite{shi2019simple}.
The resulting dataset of $218,198$ tuples of predicates and semantic arguments contains at least one Location or one Organization. Below some examples are reported: 

\begin{itemize}
    \item {\textbf{predicate}:move},{\textbf{ARG2}:to New York City}; 
    \item {\textbf{predicate}:study},{\textbf{ARGM-LOC}:in the Convent of Jesus and Mary School in New Delhi};
    \item {\textbf{predicate}:confer},{\textbf{ARG0}:by the municipality of Kautokeino and the Kautokeino Sámi Association}.
\end{itemize}

In the final step (iv), we identified the most frequently occurring combinations of `predicate,ARG0', `predicate,ARG1', and `predicate,ARG2' in order to select  a sample representative of the sentences in the data set for annotation. 

\subsection{Annotation Guidelines}
Annotation guidelines were developed in order to annotate all events in which the subject of the biography is a participant in the event. It is important to notice that there is no one-to-one correspondence between a tuple of the type <predicate,argument> and a sentence, since most sentences contain more than one predicate, as it can be observed in the following example:

``In 1974 he \underline{left} South Africa, \underline{living} in North America, Europe and the Middle East, before \underline{returning} in 1986''.

Hence, a separate annotation for each relevant subject-predicate pair was made.

The selection of the most significant semantic arguments in biographical events is guided by previous work \cite{stranisci2021mapping} in which a set of combinations of life events and named entities types were recognized as salient for biographies: locations for migrations;   organizations for education and career events. Therefore, our guidelines mainly focus on events in which the subject of the biography is involved with such named entities. Moreover, since time is a crucial feature for biographical narratives, guidelines includes the identification of temporal expressions.

\paragraph{Identification of the entity and their semantic role.} The prerequisite for an event to be annotated was that it had to involve the biography subject. This involvement was not always direct, though: an author could  be mentioned through their works, as in ``Her third novel, Missing in Machu Picchu (2013), was awarded'' or through a group they were part of, as in ``At the age of nine, her family moved to Ghana''. According to the RED guidelines\footnote{\url{https://github.com/timjogorman/RicherEventDescription}}, the former case was a BRIDGING relation, while the latter was a SET-MEMBER link. In our guidelines all these types of entity had to be annotated as if they were an instance of the writer, in order to consider important biographical events of the type `his book win a prize', in which the writer is only indirectly mentioned.     

Together with the identification of the writer, annotators had to specify her/his semantic role, in order to classify their participation in the event. Two labels were created for this purpose, both inspired by the Propbank framework: `writer-ARG0', when the entity plays roles covered by this argument, such as `Agent' or `Perceiver', `writer-ARGx', if they play roles covered by other argument types, like `Patient'. Even though grouping such arguments slightly reduces the expressiveness of the PropBank framework, it has the advantage of helping the annotators to focus on a more general distinction between events in which writers have an active role and events in which they have not.

\paragraph{Identification of events, and their taxonomy.} Events had to be annotated according to the TimeML scheme and were categorized according to a subset of TimeML event types tag: `ASP-EVENT' to mark all verbs conveying aspectual information, and `REP-EVENT', for verbs reporting other states and events, 'STATE', 'EVENT' respectively. The last two are mutually exclusive in each annotation. For instance, in the sentence ``Then, she \underline{traveled} to Venezuela, where she \underline{worked} in linguistics at the Department of Justice of Venezuela'' two separate annotations had be provided: one for the pair `she-traveled', and another one for the pair `she-worked'. `ASP-EVENT' and `REP-EVENT' may occur jointly with another `STATE' or `EVENT', in expressions such as `he started working', which results in the link structure $<started, working, ASP>$ and `he said he moved', which is encoded as $<said, moved, REP>$

Since some sentences contained nominal utterances and there were semantically empty verbs like the copular \textit{be}, guidelines allowed for the annotation of names as events or states in subordinate clauses like ``After a brief \underline{time} in Toronto'', or in nominal predicates such ``He was a \underline{professor}''. The annotation of nominal events was supported by NomBank frames \cite{meyers2004nombank}. 

\paragraph{Identification of arguments containing a location or an organization.} The third component of the guidelines was aimed at identifying the relation between the writer and some named entities that may signal their migration or their condition of being a migrant in a given place. Annotators were asked to select the entire argument containing a location or an organization, and to mark the latter as `ARGx-ORG', and the former `ARGx-LOC'. The focus of this annotation stage was not to identify the specific semantic argument, but to label the cases in which a named entity is part of a semantic role. This allowed to refine clusters of arguments and map them onto existing taxonomies. For instance, in `He works \underline{for \$organization}', the ARGx-ORG may be mapped onto the VerbNet `Beneficiary' thematic role. 

\paragraph{Identification of temporal arguments.} Finally, the guidelines establish the annotation of temporal arguments. Rather than identifying only the token triggering a time expression, the entire argument had to be selected and labelled as `ARGM-TIME'. For instance, in the example ``In 1974 he left South Africa'' the entire semantic argument ``in 1974'' had to be annotated.

A fully annotated example of the sentence below is the following:

``In 1974 he \underline{left} South Africa, \underline{living} in North America, Europe and the Middle East, before \underline{returning} in 1986''.

    $\epsilon_1$ = <he, WRITER>\\
    $\epsilon_2$ = <left, LEAVE>\\
    $\epsilon_3$ = <South Africa, LOCATION>\\
    $\epsilon_4$ = <living, LIVE>\\
    $\epsilon_5$ = <in South Africa, LOCATION>\\
    $\epsilon_6$ = <in 1974, TIME>\\
    $L_1$ = <$\epsilon_1$, $\epsilon_2$, writer-ARG0>\\
    $L_2$ = <$\epsilon_3$, $\epsilon_2$, ARGx-LOC>\\
    $L_3$ = <$\epsilon_1$, $\epsilon_4$, writer-ARG0>\\
    $L_4$ = <$\epsilon_5$, $\epsilon_4$, ARGx-LOC>\\
    $L_5$ = <$\epsilon_6$, $\epsilon_2$, ARGM-TIME>
    
\section{Annotation Task and Results}\label{sec:results}
\begin{table*}[!ht]
\begin{center}
\caption{Inter-Annotator Agreement (F-measure).\label{table:agreement}}
\resizebox{\textwidth}{!}{ 

\begin{tabular}{l|l|l|l|l|l|l|l}
annotator & Event & State & Writer-ARG0 & Writer-ARGx & ARGx-LOC & ARGx-ORG & ARGM-TIME \\
\hline
\hline
ann\_01 (baseline ann\_02) & $0.83$ & $0.72$ & $0.90$ & $0.87$ & $0.78$ & $0.75$ & $0.91$ \\
ann\_01 (baseline ann\_03) & $0.83$ & $0.76$ & $0.91$ & $0.92$ & $0.38$ & $0.75$ & $0.94$ \\
ann\_01 (baseline ann\_04) & $0.84$ & $0.66$ & $0.91$ & $0.90$ & $0.65$ & $0.83$ & $0.85$ \\
ann\_02 (baseline ann\_01) & $0.83$ & $0.66$ & $0.91$ & $0.89$ & $0.85$ & $0.94$ & $0.94$ \\
ann\_03 (baseline ann\_01) & $0.82$ & $0.64$ & $0.93$ & $0.95$ & $0.91$ & $0.92$ & $0.94$ \\
ann\_04 (baseline ann\_01) & $0.84$ & $0.61$ & $0.91$ & $0.89$ & $0.75$ & $0.70$ & $0.87$ \\
\textbf{Average} & \textbf{0.83} & \textbf{0.67} & \textbf{0.91} & \textbf{0.90} & \textbf{0.75} & \textbf{0.81} & \textbf{0.91} \\

 \end{tabular}
} 
 
 \end{center}
 \end{table*}

\begin{figure*}[!ht]
   \includegraphics[width=\textwidth]{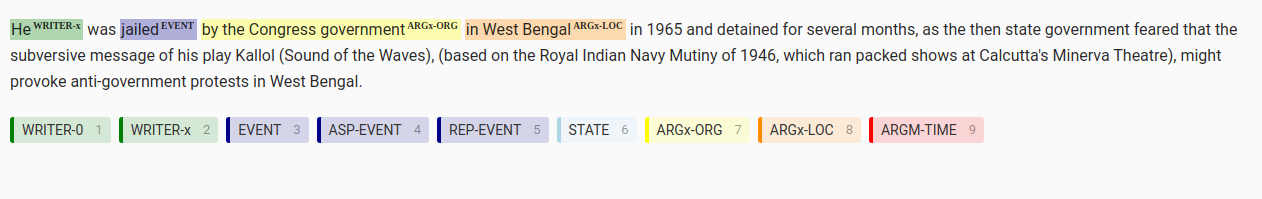}
   \includegraphics[width=\textwidth]{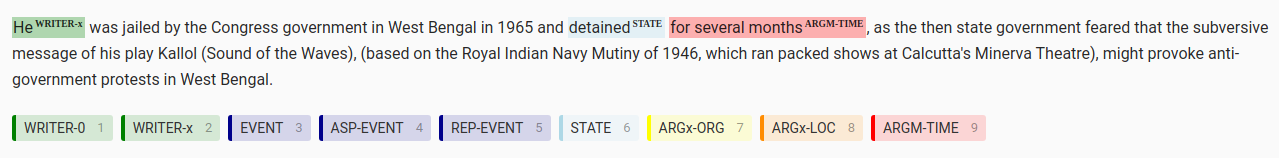}
  \caption{Two examples of annotation in Label Studio.\label{img:label_studio}}
\end{figure*}

The annotation task involved 4 annotators who evaluated $1,000$ sentences sampled from $8,047$ Wikipedia English pages of under-represented writers. One of them (ann\_01 in Table \ref{table:agreement}) evaluated all sentences $1000$, while the others annotated respectively 200 (ann\_02), 100 (ann\_03), and 200 (ann\_04) sentences. The annotation has been performed on Label Studio\footnote{\url{https://labelstud.io/}}, an Open Source platform that easily allows to organize chunk annotation tasks. Annotators were asked to provide one separate annotation for every EVENT or STATE identified in each sentence. As it is shown in Figure \ref{img:label_studio}, the same sentence has received two separated annotations. The first is the chunk `jailed' labelled as an EVENT, the second is the chunk `detained', labelled as a STATE.  

The IAA was computed through averaged pairwise F-measure: in this setting, the annotations of one annotator are used as the reference against which the annotations of the other annotator are compared. In order to maximize the agreement between annotators, we did not only consider the exact match between chunk, but also the cases in which one chunk contained the other. Adopting such an approach has allowed to resolve some recurrent inconsistencies. Let us consider the two pairs of annotations:
\begin{enumerate}
  \item awarded / was awarded \label{ex:result_1}
  \item the United Nations / to the United Nations  \label{ex:result_2}
 \end{enumerate}
 In the first one (\ref{ex:result_1}) all the smaller chunk was kept. Conversely, in (\ref{ex:result_2}) the larger chunk was kept, in order to preserve the semantic role of the argument containing an entity of the type location.
 
 Table \ref{table:agreement} shows the F-measure of the agreement between annotators for each class. Agreement is larger than 0.8 in almost all classes, with the exception of STATE and ARGx-LOC. From a qualitative analysis we observed a mismatch in the recognition of nominal events in proposition such as in (\ref{ex:result_3}). Lower agreement in ARGx-ORG identification seems to be caused by the broadness of such a type of entity that results in a variety of irrelevant usages for the annotation task, as in (\ref{ex:result_4}). 
\begin{enumerate}
  \setcounter{enumi}{2}
  \item after one year of \underline{studies} \label{ex:result_3}
  \item when Sri Lanka banned the burka on 2019, Nasrin took to \underline{Twitter} to show her support for the decision   \label{ex:result_4}
 \end{enumerate}

The resulting corpus contains $1,489$\footnote{The corpus is available at: \url{https://github.com/marcostranisci/biographicalEvents}} semantic annotations. Table \ref{tab:st_number} summarizes the number of ST in the corpus, in which there are $894$ events and $695$ states. Furthermore, $215$ aspectual or reported events were annotated; they occurred in $72$ semantic annotations. In $143$ cases, they jointly appear with an event or a state (eg: `he $[started]^{ASP-EVENT}$ $[working]^{STATE}$'). Writers hold the semantic role of agent in $1,205$ annotations, other roles in $445$. Arguments containing an organization or a location are $1,203$. More specifically, there are $281$ sentences in the corpus in which the presence of a named entity of the type LOCATION or ORGANIZATION was not relevant, despite the corpus to annotate was created by relying on a combination of Named Entity Recognition and Semantic Role Labelling (see Section \ref{sec:data_gathering}). 

\begin{table}[!bht]
 \begin{center}
 \caption{All the occurrences of Semantic Types in the corpus.
 \label{tab:st_number}}
 \begin{tabularx}{\columnwidth}{X|X}
       Semantic Type & Occurrences\\
       \hline
       \hline
EVENT & $894$ \\
STATE & $695$ \\
ASP-EVENT & $114$ \\
REP-EVENT & $101$ \\
writer-ARG0 & $1,205$ \\
writer-ARGx &	$445$ \\
ARGx-LOC & $532$	 \\
ARGx-ORG & $671$\\
TIME & $525$ \\

 \end{tabularx}
 
  \end{center}
 \end{table}

In Table \ref{tab:evaluation} the $10$ most frequently occurring events and states are shown. Some of them are related to the writers' educational journey (eg: graduate, hold, attend, study), others to their career (eg: publish, serve, teach, win, work, write). Finally, there is a set of events framing personal events (eg: bear, die, live, move). From such clusters of predicates, a set of biographical frames may be derived. This is the inverse process of existing works on biographical knowledge extraction from text \cite{menini2017ramble,russo2015extracting}. Rather than selecting a prior number of frames to be used for data gathering, this approach extracts knowledge that must subsequently be aligned to existing resources.
\begin{table}[!ht]
 \begin{center}
  \caption{The ten most frequent events and states within the corpus.}
 \begin{tabularx}{\columnwidth}{X|X|X|X}

       Event&occ.&State&occ.\\
       \hline
       \hline
receive & 56 & work & 60 \\
publish	& 39 & write & 46 \\
win &	36 & study & 41 \\
award &	34 & teach & 28 \\
write &	25 & attend & 28 \\
move &	25 & live &	21 \\
bear &	22 & serve & 20 \\
graduate & 21 & hold & 15 \\
take &	20 & spend & 14 \\
die	& 20 & writer & 13 \\

 \end{tabularx}

 \label{tab:evaluation}
  \end{center}
 \end{table}

\section{Mapping} \label{sec:mapping}
%
The annotation guidelines and the corpus presented in this paper constitute a first, yet essential step towards 
the development of a system for the automatic extraction of biographical events. While such system will be addressed in 
future work, in this Section we illustrate how the current corpus could be extended to obtain an appropriate training dataset. 
We show how the data from OntoNotes~\cite{hovy2006ontonotes} can be mapped onto our annotation schema, and report some figures regarding this process. Although OntoNotes was selected as the first target for this mapping,  the same process could also be applied to other PropBank-like datasets, such as \cite{kim2018feels}, for the enrichment of our original corpus.

%


%
OntoNotes~\cite{hovy2006ontonotes} contains a multi-layer annotation of texts from several domains (e.g., newswires, magazine articles, broadcast news). For each such domain, a PropBank-based semantic annotation and the annotation of named entities is provided. The data set is composed of $99,974$ sentences, $249,157$ rolesets, and $554,307$ semantic arguments.
Given a verb, rolesets represent all roles possibly associated to each of its senses according to the PropBank model~\cite{bonial2014propbank}.


\begin{table}[t]
\begin{center}
\caption{The distribution of arguments containing a Organization (ORG), a Person, or a Geo Political Entity (GPE) for the work.01 PropBank sense in OntoNotes.\label{arg_distribution}}
       \begin{tabular}{l|l|l|l|l}
            argument & n. & ORG & PERSON & GPE \\ 
                    \hline
                    \hline
                    ARG0 & $996$ & $6.0\%$ & $8.1\%$ & $3.5\%$ \\
                    ARG1 & $347$ & $7.8\%$ & $2.0\%$ & $7.2\%$ \\
                    ARGM-LOC & $248$ & $7.7\%$ & $0.4\%$ & $18.5\%$ \\
                    ARGM-MNR & $239$ & $0.4\%$ & $0.8\%$ & $0.0\%$ \\
                    ARGM-TMP & $148$ & $1.4\%$ & $1.4\%$ & $0\%$ \\
                    ARGM-DIS & $122$ & $0.8\%$ & $3.3\%$ & $0.0\%$ \\
                    ARG2 & $107$ & $29.0\%$ & $8.4\%$ & $11.2\%$ \\
                    ARG3 & $99$ & $17.2\%$ & $13.1\%$ & $8.1\%$ \\
            \end{tabular}
    \end{center}
\end{table}

In order to align the two corpora, we extracted all verb occurrences and their arguments. Then, we computed the percentage of arguments containing a named entity of the type ORG, GPE, or PERSON. Table~\ref{arg_distribution} shows the $8$ most frequently recurring instances for the roleset associated to work.01, which expresses the sense ``work, being employed, acts, deeds''. As it can be observed, in some of them there is a predominance of GPE and ORG compared to entities of the type PERSON. This enables the identification of some arguments that are more likely to be aligned with our corpus: it is the case of ARG1 and ARG2, which respectively correspond to `job, project' and `employer, benefactive'. 
Let us consider the following examples.

\begin{enumerate}
  \setcounter{enumi}{4}
  \item <\textit{work}, to improve China's nickel industry's level of technology, technique and equipment, ARG1>\label{ex:china_nickel}
  \item <\textit{work}, for the Justice Department, ARG2>\label{ex:justice_department}
 \end{enumerate}
 
In the former case, the GPE simply adds information about the argument, as in (\ref{ex:china_nickel}). In the latter case, it is directly linked to the verb with the role of `benefactive', as in (\ref{ex:justice_department}).

We analyzed the distribution for the $10$ most frequently occurring events and states in our corpus (Table \ref{top_10_verbs}): they amount to $430$ events, covering the $27\%$ of the overall number of instances. Besides the widespread presence of the ARGM-LOC modifier, some patterns emerge.
There is a set of events in which an ORG or a GPE has agency on the event: publish.01, award.01. The $60\%$ of the ARG0 linked to publish.01 and the $80\%$ linked to award.01 contain a GPE or a ORG. In fact, many books are \textsf{published} and many prizes are \textsf{awarded} by an organization or a geopolitical entity.

Other patterns may imply the `benefactive' role: as  mentioned before, work.01 is often linked to a benefactive as in~(\ref{ex:justice_department}). Conversely, the presence of GPE or ORG in arguments of the type `benefactive' linked to award.01 seems to be not informative. In some cases, they have an appositive function, as in `to Waring $\&$ LaRosa, New York'. At times, the mapping is less interesting for the specific task, since in some cases organizations are the recipient of a prize, which is not consistent with the biographical domain.  

Some arguments are specific to single verbs. For instance, receive.01 always presents an ARG2 associated with the role `received from', while attend.01 ARG1 always presents instances of type `thing attended'. Both combinations are common in sentences like `he attends an institution' and `he received a degree from an institution'. The distribution confirms such pattern, since the $45.1\%$ of ARG1 linked to attend.01 and the $31.7\%$ paired with receive.02 contain a ORG or a GPE.  

Finally, move.02, win.01, and work.01 show a similar behavior when an ARG1 is present. 
GPE and ORG Entities in this argument 
are not directly linked to the verb, but rather to further entities within the  argument, such as in the example~(\ref{ex:move.02}):
\begin{enumerate}
  \setcounter{enumi}{6}
  \item <`win', `the New York Drama Critics' Circle Award', ARG1>\label{ex:move.02}
\end{enumerate}
By definition, the ARG1 of the verb `win' represents a prize; however, since the organization `New York Drama Critics' is part of the argument, the entity type ORG is mistakenly considered as a value for the argument in our statistics. 
Although this behaviour represents an issue when recording descriptive statistics and for the mapping process, such dependency structures should be considered to collect precious and more subtle biographical information that needs further investigation.



Despite the actual limitations, the results of the mapping process is encouraging. In fact, even considering only non-ambiguous argument types, $851$ instances may be mapped from OntoNotes to the top ten instances of our corpus, tripling the initial size of the corpus. 
%
At the same time, we observed the emergence of patterns helpful to automatically extract and understand events and states from raw text biographies. Further studies may focus on the automatic implementation of such patterns.

\begin{table}[t]
\begin{center}
 {
 \caption{The most recurring link structures of the type <verb,argument containing ORG or GPE> for the $10$ events and states with more occurrences in our corpus.}
                  } \label{top_10_verbs}
\begin{tabular}{l|l|l}
                    
                    verb & argument(s) & description \\
                    \hline \hline
                    work.01 & ARG2 ARG1 & benefactive project \\
                    write.01 & ARG2 & benefactive \\
                    receive.01 & ARG2 & received from \\
                    publish.01 & ARG0 & publisher \\
                    win.01 & ARG1 & prize \\
                    award.01 & ARG0, ARG2 & giver, beneficiary \\
                    attend.01 & ARG1 & thing attended \\
                    move.01 & ARG2 & destination \\
                    move.02 & ARG1 & measures \\
                    study.01 & ARGM-LOC & location \\
                    teach.01 & ARGM-LOC & location \\
                    
                    \end{tabular}

                \end{center}

\end{table}


\section{Conclusions and Future Work} \label{sec:conclusions}
In this paper we presented a novel schema for the annotation of biographical events in free text. We have also built a new corpus for this task, containing $1,000$ annotated sentences sampled from $8,047$ Wikipedia English pages pertaining underrepresented writers. Finally, we have shown how existing resources, such as OntoNotes, can be mapped onto our annotation schema in order to increase significantly the size of the corpus. 

The developed corpus and the proposed schema are preparatory for the development of an automatic system for the extraction of biographical events from free text, which constitutes the main focus of our future work. Ideally, we could start from existing systems performing semantic role labeling (such as~\cite{shi2019simple}), and then adapt the results in a manner similar to the one adopted in the mapping process. The mapping process itself also needs to be strengthened with a more thorough evaluation and with the development of specific rules to better detect the entities filling the arguments.
Our final focus consists in a study aimed at better understanding and quantifying how the biographical information extracted by the system can be beneficial to tackle other downstream tasks.

\section{Bibliographical References}\label{reference}

\bibliographystyle{lrec2022-bib}
\bibliography{bibliography}

\begin{thebibliography}{}

\bibitem[\protect\citename{Allan}2012]{allan2012topic}
Allan, J.
\newblock (2012).
\newblock {\em Topic detection and tracking: event-based information
  organization}, volume~12.
\newblock Springer Science \& Business Media.

\bibitem[\protect\citename{Auer \bgroup et al.\egroup }2007]{auer2007dbpedia}
Auer, S., Bizer, C., Kobilarov, G., Lehmann, J., Cyganiak, R., and Ives, Z.
\newblock (2007).
\newblock Dbpedia: A nucleus for a web of open data.
\newblock In {\em The semantic web}, pages 722--735. Springer.

\bibitem[\protect\citename{Baker \bgroup et al.\egroup
  }1998]{baker1998berkeley}
Baker, C.~F., Fillmore, C.~J., and Lowe, J.~B.
\newblock (1998).
\newblock The berkeley framenet project.
\newblock In {\em COLING 1998 Volume 1: The 17th International Conference on
  Computational Linguistics}.

\bibitem[\protect\citename{Bonial \bgroup et al.\egroup
  }2014]{bonial2014propbank}
Bonial, C., Bonn, J., Conger, K., Hwang, J.~D., and Palmer, M.
\newblock (2014).
\newblock Propbank: Semantics of new predicate types.
\newblock In {\em Proceedings of the Ninth International Conference on Language
  Resources and Evaluation (LREC'14)}, pages 3013--3019.

\bibitem[\protect\citename{Booth}2008]{booth2008orlando}
Booth, A.
\newblock (2008).
\newblock Orlando: Women's writing in the british isles from the beginnings to
  the present.

\bibitem[\protect\citename{Bunt and Palmer}2013]{bunt2013conceptual}
Bunt, H. and Palmer, M.
\newblock (2013).
\newblock Conceptual and representational choices in defining an iso standard
  for semantic role annotation.
\newblock In {\em Proceedings Ninth Joint ISO-ACL SIGSEM Workshop on
  Interoperable Semantic Annotation (ISA-9), Potsdam}, pages 41--50.

\bibitem[\protect\citename{Bunt and Romary}2002]{bunt2002requirements}
Bunt, H. and Romary, L.
\newblock (2002).
\newblock Requirements on multimodal semantic representations.
\newblock In {\em Proceedings of ISO TC37/SC4 Preliminary Meeting}, pages
  59--68. KAIST.

\bibitem[\protect\citename{Hovy \bgroup et al.\egroup }2006]{hovy2006ontonotes}
Hovy, E., Marcus, M., Palmer, M., Ramshaw, L., and Weischedel, R.
\newblock (2006).
\newblock Ontonotes: the 90\% solution.
\newblock In {\em Proceedings of the human language technology conference of
  the NAACL, Companion Volume: Short Papers}, pages 57--60.

\bibitem[\protect\citename{Kim and Klinger}2018]{kim2018feels}
Kim, E. and Klinger, R.
\newblock (2018).
\newblock Who feels what and why? annotation of a literature corpus with
  semantic roles of emotions.
\newblock In {\em Proceedings of the 27th International Conference on
  Computational Linguistics}, pages 1345--1359.

\bibitem[\protect\citename{Kingsbury and Palmer}2002]{kingsbury2002treebank}
Kingsbury, P.~R. and Palmer, M.
\newblock (2002).
\newblock From treebank to propbank.
\newblock In {\em LREC}, pages 1989--1993. Citeseer.

\bibitem[\protect\citename{Krieger and Declerck}2015]{krieger2015owl}
Krieger, H.-U. and Declerck, T.
\newblock (2015).
\newblock An owl ontology for biographical knowledge. representing
  time-dependent factual knowledge.
\newblock In {\em BD}, pages 101--110.

\bibitem[\protect\citename{Manning \bgroup et al.\egroup
  }2014]{manning2014stanford}
Manning, C.~D., Surdeanu, M., Bauer, J., Finkel, J.~R., Bethard, S., and
  McClosky, D.
\newblock (2014).
\newblock The stanford corenlp natural language processing toolkit.
\newblock In {\em Proceedings of 52nd annual meeting of the association for
  computational linguistics: system demonstrations}, pages 55--60.

\bibitem[\protect\citename{Menini \bgroup et al.\egroup
  }2017]{menini2017ramble}
Menini, S., Sprugnoli, R., Moretti, G., Bignotti, E., Tonelli, S., and Lepri,
  B.
\newblock (2017).
\newblock Ramble on: Tracing movements of popular historical figures.
\newblock In {\em Proceedings of the Software Demonstrations of the 15th
  Conference of the European Chapter of the Association for Computational
  Linguistics}, pages 77--80.

\bibitem[\protect\citename{Meyers \bgroup et al.\egroup
  }2004]{meyers2004nombank}
Meyers, A., Reeves, R., Macleod, C., Szekely, R., Zielinska, V., Young, B., and
  Grishman, R.
\newblock (2004).
\newblock The nombank project: An interim report.
\newblock In {\em Proceedings of the workshop frontiers in corpus annotation at
  hlt-naacl 2004}, pages 24--31.

\bibitem[\protect\citename{O’Gorman \bgroup et al.\egroup }2016]{o2016richer}
O’Gorman, T., Wright-Bettner, K., and Palmer, M.
\newblock (2016).
\newblock Richer event description: Integrating event coreference with
  temporal, causal and bridging annotation.
\newblock In {\em Proceedings of the 2nd Workshop on Computing News Storylines
  (CNS 2016)}, pages 47--56.

\bibitem[\protect\citename{Petukhova and Bunt}2008]{petukhova-bunt-2008-lirics}
Petukhova, V. and Bunt, H.
\newblock (2008).
\newblock {LIRICS} semantic role annotation: Design and evaluation of a set of
  data categories.
\newblock In {\em Proceedings of the Sixth International Conference on Language
  Resources and Evaluation ({LREC}'08)}, Marrakech, Morocco, May. European
  Language Resources Association (ELRA).

\bibitem[\protect\citename{Pustejovsky \bgroup et al.\egroup
  }2010]{pustejovsky2010iso}
Pustejovsky, J., Lee, K., Bunt, H., and Romary, L.
\newblock (2010).
\newblock Iso-timeml: An international standard for semantic annotation.
\newblock In {\em LREC}, volume~10, pages 394--397.

\bibitem[\protect\citename{Russo \bgroup et al.\egroup
  }2015]{russo2015extracting}
Russo, I., Caselli, T., and Monachini, M.
\newblock (2015).
\newblock Extracting and visualising biographical events from wikipedia.
\newblock In {\em BD}, pages 111--115.

\bibitem[\protect\citename{Schuler}2005]{schuler2005verbnet}
Schuler, K.~K.
\newblock (2005).
\newblock {\em VerbNet: A broad-coverage, comprehensive verb lexicon}.
\newblock University of Pennsylvania.

\bibitem[\protect\citename{Shi and Lin}2019]{shi2019simple}
Shi, P. and Lin, J.
\newblock (2019).
\newblock Simple bert models for relation extraction and semantic role
  labeling.
\newblock {\em arXiv preprint arXiv:1904.05255}.

\bibitem[\protect\citename{Song \bgroup et al.\egroup }2015]{song2015light}
Song, Z., Bies, A., Strassel, S.~M., Riese, T., Mott, J., Ellis, J., Wright,
  J., Kulick, S., Ryant, N., Ma, X., et~al.
\newblock (2015).
\newblock From light to rich ere: Annotation of entities, relations, and
  events.
\newblock In {\em EVENTS@ HLP-NAACL}, pages 89--98.

\bibitem[\protect\citename{Stranisci \bgroup et al.\egroup
  }2021a]{stranisci2021mapping}
Stranisci, M.~A., Basile, V., Damiano, R., and Patti, V.
\newblock (2021a).
\newblock Mapping biographical events to odps through lexico-semantic patterns?
\newblock In {\em 12th Workshop on Ontology Design and Patterns, WOP 2021},
  volume 3011, pages 1--12. CEUR-WS.

\bibitem[\protect\citename{Stranisci \bgroup et al.\egroup
  }2021b]{stranisci2021representing}
Stranisci, M.~A., Patti, V., and Damiano, R.
\newblock (2021b).
\newblock Representing the under-represented: A dataset of post-colonial, and
  migrant writers.
\newblock In {\em 3rd Conference on Language, Data and Knowledge, LDK 2021},
  volume~93, pages 1--14. Schloss Dagstuhl-Leibniz-Zentrum fur Informatik GmbH,
  Dagstuhl Publishing.

\bibitem[\protect\citename{Sun and Peng}2021]{sun2021men}
Sun, J. and Peng, N.
\newblock (2021).
\newblock Men are elected, women are married: Events gender bias on wikipedia.
\newblock In {\em Proceedings of the 59th Annual Meeting of the Association for
  Computational Linguistics and the 11th International Joint Conference on
  Natural Language Processing (Volume 2: Short Papers)}, pages 350--360.

\bibitem[\protect\citename{Tuominen \bgroup et al.\egroup
  }2018]{tuominen2018bio}
Tuominen, J.~A., Hyv{\"o}nen, E.~A., Leskinen, P., et~al.
\newblock (2018).
\newblock Bio crm: A data model for representing biographical data for
  prosopographical research.
\newblock In {\em Proceedings of the Second Conference on Biographical Data in
  a Digital World 2017 (BD2017)}. CEUR Workshop Proceedings.

\bibitem[\protect\citename{Vrande{\v{c}}i{\'c} and
  Kr{\"o}tzsch}2014]{vrandevcic2014wikidata}
Vrande{\v{c}}i{\'c}, D. and Kr{\"o}tzsch, M.
\newblock (2014).
\newblock Wikidata: a free collaborative knowledgebase.
\newblock {\em Communications of the ACM}, 57(10):78--85.

\bibitem[\protect\citename{Wang \bgroup et al.\egroup }2018]{wang2018acekg}
Wang, R., Yan, Y., Wang, J., Jia, Y., Zhang, Y., Zhang, W., and Wang, X.
\newblock (2018).
\newblock Acekg: A large-scale knowledge graph for academic data mining.
\newblock In {\em Proceedings of the 27th ACM international conference on
  information and knowledge management}, pages 1487--1490.

\bibitem[\protect\citename{Xiang and Wang}2019]{xiang2019survey}
Xiang, W. and Wang, B.
\newblock (2019).
\newblock A survey of event extraction from text.
\newblock {\em IEEE Access}, 7:173111--173137.

\end{thebibliography}


\begin{thebibliography}{}

\bibitem[\protect\citename{Pustejovsky \bgroup et al.\egroup
  }2003]{pustejovsky2003timebank}
Pustejovsky, J., Hanks, P., Sauri, R., See, A., Gaizauskas, R., Setzer, A.,
  Radev, D., Sundheim, B., Day, D., Ferro, L., et~al.
\newblock (2003).
\newblock The timebank corpus.
\newblock In {\em Corpus linguistics}, volume 2003, page~40. Lancaster, UK.

\end{thebibliography}

\section{Language Resource References}
\label{lr:ref}
\bibliographystylelanguageresource{lrec2022-bib}
\bibliographylanguageresource{languageresource}

\end{document}